\pdfoutput=1
\documentclass{article}

    \usepackage[preprint]{neurips}

\usepackage[utf8]{inputenc} 
\usepackage[T1]{fontenc}    
\usepackage{url}            
\usepackage{booktabs}       
\usepackage{algorithm}
\usepackage{algorithmic}
\usepackage{amsmath}
\usepackage{amssymb}
\usepackage{mathtools}
\usepackage{amsthm}
\usepackage{amsfonts}       
\usepackage{nicefrac}       
\usepackage{microtype}      
\usepackage{xcolor}         

\usepackage{paralist}
\newtheorem{thm}{Theorem}
\newtheorem{lem}{Lemma}
\newtheorem{myDef}{Definition}
\newtheorem{assum}{Assumption}

\def \ze {\mathbf{0}}
\def \A {\mathcal{A}}

\def \K {\mathcal{K}}

\def \B {\mathcal{B}}

\def \F {\mathcal{F}}
\def \E {\mathbb{E}}
\def \x {\mathbf{x}}
\def \y {\mathbf{y}}

\def \g {\mathbf{g}}

\def \v {\mathbf{v}}
\def \u {\mathbf{u}}
\def \y {\mathbf{y}}

\DeclareMathOperator*{\re}{Reg}

\DeclareMathOperator*{\argmin}{argmin}

\title{Improved Regret for Bandit Convex Optimization\\ with Delayed Feedback}

\author{%
  Yuanyu Wan\textsuperscript{\rm 1},~~~Chang Yao\textsuperscript{\rm 1},~~~Mingli Song\textsuperscript{\rm 1},~~~Lijun Zhang\textsuperscript{\rm 2}\\
  \textsuperscript{\rm 1}State Key Laboratory of Blockchain and Data Security, Zhejiang University, Hangzhou, China\\
  \textsuperscript{\rm 2}National Key Laboratory for Novel Software Technology, Nanjing University, Nanjing, China\\
  \texttt{\{wanyy,changy,brooksong\}@zju.edu.cn,}~~\texttt{zhanglj@lamda.nju.edu.cn}
}

\begin{document}

\maketitle

\begin{abstract}
We investigate bandit convex optimization (BCO) with delayed feedback, where only the loss value of the action is revealed under an arbitrary delay. Let $n,T,\bar{d}$ denote the dimensionality, time horizon, and average delay, respectively. Previous studies have achieved an $O(\sqrt{n}T^{3/4}+(n\bar{d})^{1/3}T^{2/3})$ regret bound for this problem, whose delay-independent part matches the regret of the classical non-delayed bandit gradient descent algorithm. However, there is a large gap between its delay-dependent part, i.e., $O((n\bar{d})^{1/3}T^{2/3})$, and an existing $\Omega(\sqrt{\bar{d}T})$ lower bound. In this paper, we illustrate that this gap can be filled in the worst case, where $\bar{d}$ is very close to the maximum delay $d$. Specifically, we first develop a novel algorithm, and prove that it enjoys a regret bound of $O(\sqrt{n}T^{3/4}+\sqrt{dT})$ in general. Compared with the previous result, our regret bound is better for $d=O((n\bar{d})^{2/3}T^{1/3})$, and the delay-dependent part is tight in the worst case. The primary idea is to decouple the joint effect of the delays and the bandit feedback on the regret by carefully incorporating the delayed bandit feedback with a blocking update mechanism. Furthermore, we show that the proposed algorithm can improve the regret bound to $O((nT)^{2/3}\log^{1/3}T+d\log T)$ for strongly convex functions. Finally, if the action sets are unconstrained, we demonstrate that it can be simply extended to achieve an $O(n\sqrt{T\log T}+d\log T)$ regret bound for strongly convex and smooth functions.
\end{abstract}

\section{Introduction}
Online convex optimization (OCO) with delayed feedback \citep{Joulani13,Quanrud15} has become a popular paradigm for modeling streaming applications without immediate reactions to actions, such as online advertisement \citep{McMahan2013} and online routing \citep{Awerbuch2008}. Formally, it is defined as a repeated game between a player and an adversary. At each round $t$, the player first selects an action $\x_t$ from a convex set $\K\subseteq\mathbb{R}^n$. Then, the adversary chooses a convex function $f_t(\cdot):\mathbb{R}^n\mapsto\mathbb{R}$, which causes the player a loss $f_t(\x_t)$ but is revealed at the end of round $t+d_t-1$, where $d_t\geq 1$ denotes an arbitrary delay. The goal of the player is to minimize the regret, i.e., the gap between the cumulative loss of the player and that of an optimal fixed action
\[\re(T)=\sum_{t=1}^Tf_t(\x_t)-\min_{\x\in\K}\sum_{t=1}^Tf_t(\x)\]
where $T$ is the number of total rounds.

Over the past decades, plenty of algorithms and theoretical guarantees have been proposed for this problem \citep{Weinberger02_TIT,Langford09,Joulani13,Quanrud15,Joulani16,ICML20_Mertikopoulos,ICML21_delay,DOGD-SC,NeurIPS22-Wan,Bistritz-JMLR22}. However, the vast majority of them assume that the full information or gradients of delayed functions are available for updating the action, which is not necessarily satisfied in reality. For example, in online routing \citep{Awerbuch2008}, the player selects a path through a given network for some packet, and its loss is measured by the time length of the path. Although this loss value can be observed after the packet arrives at the destination, the player rarely has access to the congestion pattern of the entire network \citep{Hazan2016}. To address this limitation, it is natural to investigate a more challenging setting, namely bandit convex optimization (BCO) with delayed feedback, where only the loss value $f_t(\x_t)$ is revealed at the end of round $t+d_t-1$.

It is well known that in the non-delayed BCO, bandit gradient descent (BGD), which performs the gradient descent step based on a one-point estimator of the gradient, enjoys a regret bound of $O(\sqrt{n}T^{3/4})$ \citep{OBO05}. Despite its simplicity, without additional assumptions on functions, there does not exist any practical algorithm that can improve the regret of BGD.~Therefore, a few studies have proposed to extend BGD and its regret bound into the delayed setting \citep{ICML20_Mertikopoulos,Bistritz-JMLR22}. Specifically, \citet{ICML20_Mertikopoulos} first propose an algorithm called gradient-free online learning with delayed feedback (GOLD), which utilizes the oldest received but not utilized loss value to perform an update similar to BGD at each round. Let $d=\max\{d_1,\dots,d_T\}$ denote the maximum delay. According to the analysis of \citet{ICML20_Mertikopoulos}, GOLD can achieve a regret bound of $O(\sqrt{n}T^{3/4}+(nd)^{1/3}T^{2/3})$, which matches the $O(\sqrt{n}T^{3/4})$ regret of BGD in the non-delayed setting for $d=O(\sqrt{n}T^{1/4})$. Very recently, \citet{Bistritz-JMLR22} develop an improved variant of GOLD by utilizing all received but not utilized loss values one by one at each round, and reduce the regret bound to $O(\sqrt{n}T^{3/4}+(n\bar{d})^{1/3}T^{2/3})$,\footnote{Note that \citet{Bistritz-JMLR22} actually only argue a regret bound of $O(nT^{3/4}+\sqrt{n}\bar{d}^{1/3}T^{2/3})$. However, as discussed in our Appendix \ref{app1}, it is not hard to derive this refined bound by tuning parameters more carefully.} where $\bar{d}=(1/T)\sum_{t=1}^Td_t$ is the average delay. However, there still exists a large gap between the delay-dependent part in the improved bound and an existing $\Omega(\sqrt{\bar{d}T})$ lower bound \citep{Bistritz-JMLR22}. It remains unclear whether this gap can be filled, especially by improving the existing upper bound.

In this paper, we provide an affirmative answer to this question in the worst case, where $\bar{d}$ is very close to $d$. Specifically, we first develop a new algorithm, namely delayed follow-the-bandit-leader (D-FTBL), and show that it enjoys a regret bound of $O(\sqrt{n}T^{3/4}+\sqrt{dT})$ in general. Notice that both the $O((nd)^{1/3}T^{2/3})$ and $O((n\bar{d})^{1/3}T^{2/3})$ terms in previous regret bounds \citep{ICML20_Mertikopoulos,Bistritz-JMLR22} can be attributed to the joint effect of the delays, and the one-point gradient estimator, especially its large variance depending on the exploration radius. To improve the regret, besides the one-point gradient estimator, we further incorporate the delayed bandit feedback with a blocking update mechanism, i.e., dividing total $T$ rounds into several equally-sized blocks and only updating the action at the end of each block. Despite its simplicity, there exist two nice properties about the cumulative estimated gradients at each block.
\begin{compactitem}
  \item First, with an appropriate block size, its variance becomes proportional to only the block size without extra dependence on the exploration radius.
  \item Second, the block-level delay suffered by the cumulative estimated gradients at each block is in reverse proportion to the block size.
\end{compactitem}
Surprisingly, by combining these properties, the previous joint effect of the delays and the one-point gradient estimator can be decoupled, which is critical for deriving our regret bound. 
Compared with the existing results, in the worst case, our regret bound matches the $O(\sqrt{n}T^{3/4})$ regret of the non-delayed BGD for a larger amount of delays, i.e., $d=O(n\sqrt{T})$, and the delay-dependent part, i.e., $O(\sqrt{dT})$, matches the lower bound. Moreover, it is worth noting that our regret bound actually is better than that of \citet{Bistritz-JMLR22} as long as $d$ is not larger than $O((n\bar{d})^{2/3}T^{1/3})$, which even covers the case with $\bar{d}=O(1)$ partially. To the best of our knowledge, this is the first work that shows the benefit of the blocking update mechanism in delayed BCO, though it is commonly utilized to develop projection-free algorithms for efficiently dealing with complicated action sets \citep{MingruiZhang-NIPS19,Garber19,Hazan20,Wan-ICML-2020}.

Furthermore, we consider the special case of delayed BCO with strongly convex functions. In the non-delayed setting, \citet{Agarwal2010_COLT} have shown that BGD can improve the regret from $O(\sqrt{n}T^{3/4})$ to $O((nT)^{2/3}\log^{1/3}T)$ by exploiting the strong convexity. If functions are also smooth and the action set is unconstrained, BGD has been  extended to achieve an $O(n\sqrt{T\log T})$ regret bound \citep{Agarwal2010_COLT}. Analogous to these improvements, we prove that our D-FTBL can achieve a regret bound of $O((nT)^{2/3}\log^{1/3}T+d\log T)$ for strongly convex functions, and its simple extension enjoys a regret bound of $O(n\sqrt{T\log T}+d\log T)$ for strongly convex and smooth functions over unconstrained action sets. These regret bounds also match those of BGD in the non-delayed setting for a relatively large amount of delay. Moreover, the $O(d\log T)$ part in these two bounds matches an $\Omega(d\log T)$ lower bound adapted from the easier full-information setting with strongly convex and smooth functions \citep{Weinberger02_TIT}.





\section{Related work}
In this section, we briefly review the related work on online convex optimization (OCO) and bandit convex optimization (BCO), as well as delayed feedback.

\subsection{Standard OCO and BCO}
If $d_t=1$ for all $t\in[T]$, OCO with delayed feedback reduces to the standard OCO \citep{Zinkevich2003}. Online gradient descent (OGD) \citep{Zinkevich2003,Hazan_2007} is one of the most popular algorithm for this problem, which simply updates the action $\x_t$ via a gradient descent step based on $\nabla f_t(\x_t)$.~By using appropriate step sizes, OGD can achieve $O(\sqrt{T})$ and $O(\log T)$ regret bounds for convex and strongly convex functions, respectively. Follow-the-regularized-leader (FTRL) \citep{Hazan_2007,Online:suvery,Hazan2016} is an alternative algorithm, which chooses the new action by minimizing the linear approximation of cumulative loss functions under some regularization. With appropriate regularization, FTRL achieves the same $O(\sqrt{T})$ and $O(\log T)$ regret bounds as OGD. Moreover, \citet{Abernethy08} have presented a lower bound of $\Omega(\sqrt{T})$ for convex functions, and a refined lower bound of $\Omega(\log T)$ for strongly convex functions, which implies that both OGD and FTRL are optimal. 

BCO is a special yet more challenging case of OCO, where the player can only receive the loss value $f_t(\x_t)$ at each round $t$. The first algorithm for BCO is bandit gradient descent (BGD) \citep{OBO05}, which replaces the exact gradient used in OGD with an estimated gradient based on the single loss value (known as the classical one-point gradient estimator). By incorporating the approximation error of gradients into the regret analysis of OGD, \citet{OBO05} establish an $O(\sqrt{n}T^{3/4})$ regret bound for BGD with convex functions. Later, \citet{Agarwal2010_COLT} show that BGD enjoys an $O((nT)^{2/3}\log^{1/3}T)$ regret bound for strongly convex functions, and can be extended to achieve an $O(n\sqrt{T\log T})$ regret bound in the special case of unconstrained BCO with strongly convex and smooth functions. \citet{Saha_2011} develop a new algorithm for BCO with smooth functions, and establish the $O((nT)^{2/3}\log^{1/3}T)$ regret bound without the strongly convex assumption. \citet{Hoeven-NIPS20} propose novel BCO algorithms, which adaptively improve the previous regret bounds for convex and smooth functions if the norm of the comparator is small. By revisiting the case with strongly convex and smooth functions, several algorithms \citep{BCO_Hazan,Shinji_2020} have been developed to achieve the $O(n\sqrt{T\log T})$ regret bound in the constrained setting. 

Moreover, a series of studies \citep{Bubeck-COLT16,Hazan-2016-arxiv,Bubeck_2017,Lattimore-BCO-20,Bubeck-JOA21} have been devoted to designing nearly optimal algorithms, which almost match the $\Omega(n\sqrt{T})$ lower bound for the general BCO \citep{Shamir_2013} without any additional assumption. However, the running time of their algorithms are either exponential in $n$ and $T$, or polynomial with a high degree on $n$ and $T$, which is not suitable for practical large-scale applications. We refer the interested reader to \citet{BCO_survey24} for a comprehensive survey on BCO. Additionally, we notice that BCO is closely related to the zero-order stochastic optimization (ZOSO) problem \citep{Duchi_ZeroOrder,Bach_COLT16,Shamir_Zero_JMLR}, where the stochastic values are available for minimizing a fixed loss function. However, ZOSO is less challenging than BCO in the sense that it does not need to deal with time-varying functions and is usually allowed to query the loss value at two points per iteration.

\subsection{OCO and BCO with delays}
The seminal work of \citet{Weinberger02_TIT} first considers the case with a fixed delay, i.e., $d_t=d$ for all $t\in[T]$, and proposes a black-box technique that can covert any traditional OCO algorithm into the delayed setting. The main idea is to maintain $d$ instances of the traditional algorithm, and alternately utilize these instances to generate the new action.~If the regret of the traditional algorithm is bounded by $\re(T)$, this technique can achieve an $d\re(T/d)$ regret bound.~Moreover, there exist $\Omega(\sqrt{dT})$ and $\Omega(d\log T)$ lower bounds for convex functions, and strongly convex and smooth functions, respectively \citep{Weinberger02_TIT}. However, the delays are not always fixed in practice,  and its space complexity is $d$ times as much as that of the traditional algorithm, which could be prohibitively resource-intensive.~Although \citet{Joulani13} have generalized this technique to deal with arbitrary delays, the space complexity remains high. Besides these black-box techniques, there exists a surge of interest in developing and analyzing specialized algorithms for delayed OCO \citep{Langford09,McMahan14,Quanrud15,Joulani16,Li_AISTATS19,ICML21_delay,DOGD-SC,NeurIPS22-Wan,Wan-Arxiv-22}, which either do not require additional computational resources or enjoy better regret bounds. 

Despite the great flourish of research on OCO with delays and BCO, delayed BCO has rarely been investigated.~GOLD \citep{ICML20_Mertikopoulos} is the first algorithm for this problem, which originally has the $O(\sqrt{n}T^{3/4}+(nd)^{1/3}T^{2/3})$ regret, and is further refined to enjoy the $O(\sqrt{n}T^{3/4}+(n\bar{d})^{1/3}T^{2/3})$ regret \citep{Bistritz-JMLR22}. However, \citet{Bistritz-JMLR22} also present an unmatched lower bound of $\Omega(\sqrt{\bar{d}T})$. Although two recent advances in a more complicated bandit non-stochastic control problem \citep{Gradu-NIPS20,Sun-NIPS-23} provide some intermediate results about OGD and FTRL with the delayed bandit feedback, they focus on the case with a fixed delay and can only recover the $O(\sqrt{n}T^{3/4}+(nd)^{1/3}T^{2/3})$ regret in general. In this paper, we take one further step toward understanding the effect of arbitrary delays on BCO by establishing improved regret bounds such that the delay-independent part is equal to the regret of BGD, and the delay-dependent part matches the lower bound in the worst case. Moreover, we notice that although the block-box technique of \citet{Joulani13} can also convert BGD into the delayed setting, it only achieves an $O(\sqrt{n}d^{1/4}T^{3/4})$ regret bound for convex functions, which is much worse than that of GOLD and our algorithm.

\section{Main results}
In this section, we first introduce the necessary preliminaries including definitions, assumptions, and an algorithmic ingredient. Then, we present our improved algorithm for BCO with delayed feedback, as well as the corresponding theoretical guarantees. 

\subsection{Preliminaries}
\label{sec-Pre}
We first recall two standard definitions about the smoothness and strong convexity of functions \citep{Boyd04}.
\begin{myDef}
\label{def1}
A function $f(\x):\mathbb{R}^n\to\mathbb{R}$ is called $\beta$-smooth over $\K$ if for all $\x,\mathbf{y}\in \K$, it holds that $f(\mathbf{y})\leq f(\x)+\langle\nabla f(\x),\mathbf{y}-\x\rangle+\frac{\beta}{2}\|\mathbf{y}-\x\|_2^2$.
\end{myDef}
\begin{myDef}
\label{def0}
A function $f(\x):\mathbb{R}^n\to\mathbb{R}$ is called $\alpha$-strongly convex over $\K$ if for all $\x,\y\in\K$, it holds that $f(\mathbf{y})\geq f(\x)+\langle\nabla f(\x),\mathbf{y}-\x\rangle+\frac{\alpha}{2}\|\mathbf{y}-\x\|_2^2$.
\end{myDef}

Note that as proved by \citet{Hazan2012}, any $\alpha$-strongly convex function $f(\x):\mathbb{R}^n\mapsto\mathbb{R}$ over the convex set $\K$ ensures that
\begin{equation}
\label{strong-nice-pro}
\frac{\alpha}{2}\left\|\x-\x^\ast\right\|_2^2\leq f(\x)-f(\x^\ast)
\end{equation}
for any $\x\in\K$, where $\x^\ast=\argmin_{\x\in\K}f(\x)$.

Then, following previous studies on BCO \citep{OBO05,ICML20_Mertikopoulos,Garber19,Garber-AISTATS21}, we introduce some common assumptions.
\begin{assum}
\label{assum3}
The convex set $\K$ is full-dimensional and contains the origin, and there exist two constants $r,R>0$ such that $r\B^n\subseteq \K\subseteq R\B^n$, where $\B^n$ denotes the unit Euclidean ball centered at the origin in $\mathbb{R}^n$.
\end{assum}
\begin{assum}
\label{assum1}
All loss functions are $G$-Lipschitz over $\K$, i.e., for all $\x,\y\in\K$ and $t\in[T]$, it holds that $|f_t(\x)-f_t(\y)|\leq G\|\x-\y\|_2$.
\end{assum}
\begin{assum}
\label{assum2}
The absolute value of all loss functions over $\K$ are bounded by $M$, i.e., for all $\x\in\K$ and $t\in[T]$, it holds that $|f_t(\x)|\leq M$. Additionally, all loss functions are chosen beforehand, i.e.,
the adversary is oblivious.
\end{assum}
Finally, we introduce the one-point gradient estimator \citep{OBO05}, which is a standard technique for exploiting the bandit feedback. Given a function $f(\x):\mathbb{R}^n\mapsto\mathbb{R}$, we can define the $\delta$-smoothed version of $f(\x)$ as
\begin{equation}
\label{delta-smooth}
\hat{f}_\delta(\x)=\mathbb{E}_{\u\sim\B^n}[f(\x+\delta\u)]
\end{equation}
where the parameter $\delta\in(0,1)$ is the so-called exploration radius. As proved by \citet{OBO05}, the $\delta$-smoothed version satisfies the following lemma.
\begin{lem}
\label{smoothed_lem2}
(Lemma 1 in \citet{OBO05})
Given a function $f(\x):\mathbb{R}^n\mapsto\mathbb{R}$ and a constant $\delta\in(0,1)$, its $\delta$-smoothed version $\hat{f}_\delta(\x)$ defined in \eqref{delta-smooth} ensures
\[\nabla\hat{f}_\delta(\x)=\mathbb{E}_{\u\sim\mathcal{S}^n}\left[\frac{n}{\delta}f(\x+\delta\u)\u\right]\]
where $\mathcal{S}^n$ denotes the unit Euclidean sphere centered at the origin in $\mathbb{R}^n$.
\end{lem}
From Lemma \ref{smoothed_lem2}, the randomized vector $\frac{n}{\delta}f(\x+\delta\u)\u$, which can be computed by only utilizing a single loss value, is an unbiased estimator of $\nabla\hat{f}_\delta(\x)$. Moreover, \citet{OBO05} have also shown that $\hat{f}_\delta(\x)$ is close to the original function $f(\x)$ over a shrunk set 
\begin{equation}
\label{shrink-set}
\K_\delta=(1-\delta/r)\K=\{(1-\delta/r)\x|\x\in\K\}.
\end{equation}
Therefore, this one-point estimator can be utilized as a good substitute for the gradient $\nabla f(\x)$ in the bandit setting. For example, we notice that at each round $t$, BGD \citep{OBO05} first plays an action $\x_t=\y_t+\u_t$, where $\y_t\in\K_\delta$ and $\u_t\sim\mathcal{S}^n$, and then updates $\y_t$ as
\begin{equation}
\label{BGD-up}
\y_{t+1}=\Pi_{\K_\delta}\left(\y_t-\frac{\eta_t n}{\delta}f_t(\x_t)\u_t\right)
\end{equation}
where $\Pi_{\K_\delta}(\y)=\argmin_{\x\in\K_\delta}\|\x-\y\|_2^2$ denotes the projection onto the set $\K_\delta$, and $\eta_t$ is the step size.




\subsection{Our improved algorithm}
Before introducing our algorithm, we first briefly discuss the joint effect of the delays and the bandit feedback in GOLD \citep{ICML20_Mertikopoulos},
which will provide insights for our improvements. Recall that in the delayed setting, the loss value $f_t(\x_t)$ will be delayed to the end of round $t+d_t-1$, and thus the player can only receive $\{f_k(\x_k)|k\in\F_t\}$ at the end of round $t$, where $\F_t=\{k|k+d_k-1=t\}$. Since the set $\F_t$ may not contain the round $t$, the vanilla BGD in \eqref{BGD-up} is no longer valid. To address this issue, GOLD \citep{ICML20_Mertikopoulos} replaces $f_t(\x_t)$ in \eqref{BGD-up} with the oldest received but not utilized loss value at the end of round $t$. Intuitively, the update of this approach is $O(d)$ rounds slower than that of the vanilla BGD, which is analogous to those delayed OCO algorithms. However, due to the use of the one-point gradient estimator, the slower update causes a difference of $O(\eta dn/\delta)$ between its action and that of BGD, and the cumulative difference will bring additional regret of $O(T\eta dn/\delta)$, where a constant step size $\eta_t=\eta$ is discussed for brevity. Note that from the standard analysis of BGD, to control the total exploration cost, the value of $1/\delta$ should be sublinear in $T$. Therefore, it will amplify the effect of delays, and finally results in the $O(\sqrt{n}T^{3/4}+(nd)^{1/3}T^{2/3})$ regret \citep{ICML20_Mertikopoulos}.

To reduce the effect of delays, we propose to incorporate the delayed bandit feedback with a blocking update mechanism \citep{MingruiZhang-NIPS19,Garber19}. Specifically, we divide the total $T$ rounds into $T/K$ blocks, each with $K$ rounds, where $T/K$ is assumed to be an integer without loss of generality. For each block $m\in[T/K]$, we only maintain a preparatory action $\y_m\in\K_\delta$, and play $\x_t=\y_m+\delta\u_t$ with $\u_t\sim\mathcal{S}^n$ at each round $t$ in the block. Due to the randomness of $\u_t$ and the independence of $\x_t$ in the same block, it is not hard to verify that for each block $m\in[T/K]$, the sum of randomized gradients generated by the one-point estimator, i.e.,
$\nabla_m=\sum_{t=(m-1)K+1}^{mK}\frac{n}{\delta}f_t(\x_t)\u_t$ satisfies (see Lemma \ref{block-gradient} presented in Section \ref{sec-4.1} for details) 
\[\E[\|\nabla_m\|_2]=O(\sqrt{Kn^2/\delta^2}+K).\] By using an appropriate block size of $K=O(n^2/\delta^2)$, this upper bound will be $\E[\|\nabla_m\|_2]=O(K)$. By contrast, without the blocking update mechanism, one can only achieve $\E[\|\nabla_m\|_2]=O(Kn/\delta)$. Moreover, we notice that the cumulative estimated gradients $\nabla_m$ will be delayed at most $O(d/K)$ blocks, because even the last component $\frac{n}{\delta}f_{mK}(\x_{mK})\u_{mK}$ is available at the end of round $mK+d-1$. 

As a result, one possible approach to determine $\y_m$ for each block is to extend the update rule of GOLD \citep{ICML20_Mertikopoulos} into the block level with $K=O(n^2/\delta^2)$. Combining with previous discussions, it will reduce the effect of delays on the regret from $O(T\eta dn/\delta)$ to
\[
O\left(T\eta\frac{d}{K}\left(\sqrt{\frac{Kn^2}{\delta^2}}+K\right)\right){=}O\left(\eta dT\right)
\]
which is good enough for deriving our desired regret bounds. However, it requires a bit complicated procedure to maintain the cumulative estimated gradients for any block that has not been utilized to update the action. For this reason, instead of utilizing this approach, we incorporate FTRL \citep{Hazan_2007,Hazan2016} with the delayed bandit feedback and blocking update mechanism, which provides a more elegant way to utilize the delayed information.
\begin{algorithm}[t]
\caption{Delayed Follow-The-Bandit-Leader}
\label{alg1}
\begin{algorithmic}[1]
\STATE \textbf{Input:} $\delta, K,\alpha$, and $\eta>0$ if $\alpha=0$
\STATE \textbf{Initialization:} set $\bar{\g}_0=\ze$ and choose $\y_1\in \K_\delta$ arbitrarily
\FOR{$m=1,2,\dots,T/K$}
\FOR{$t=(m-1)K+1,\dots,mK$}
\STATE Play $\x_t=\y_{m}+\delta \u_t$, where $\u_t\sim\mathcal{S}^n$
\STATE Query $f_t(\x_t)$, and receive $\{f_k(\x_k)|k\in \F_t\}$
\STATE Update $\bar{\g}_t=\bar{\g}_{t-1}+\sum_{k\in\F_t}\frac{n}{\delta}f_k(\x_k)\u_k$
\ENDFOR
\STATE Set $
\mathcal{R}_m(\x)=\left\{\begin{array}{lr}
\frac{1}{\eta}\|\x-\y_1\|_2^2 & \text{ if }\alpha=0\\
\sum_{i=1}^{m}\frac{K\alpha}{2}\|\x-\y_i\|_2^2 & \text{otherwise}
\end{array}
\right.
$
\STATE $\y_{m+1}=\argmin_{\x\in\K_\delta}\left\{\langle\bar{\g}_{mK},\x\rangle+\mathcal{R}_m(\x)\right\}$
\ENDFOR
\end{algorithmic}
\end{algorithm}

Specifically, we initialize $\y_1\in\K_\delta$ arbitrarily, and use a variable $\bar{\g}_t$ to record the sum of gradients estimated from all received loss values, i.e., $\bar{\g}_t=\sum_{i=1}^t\sum_{k\in\F_i}\frac{n}{\delta}f_k(\x_k)\u_k$. According to FTRL, an ideal $\y_{m+1}$ should be selected by minimizing the linear approximation of cumulative loss functions under some regularization, i.e.,
\begin{equation}
\label{Ideal-FTRL}
\y_{m+1}=\argmin_{\x\in\K_\delta}\left\{\sum_{i=1}^{m}\left\langle\nabla_i,\x\right\rangle+\mathcal{R}_m(\x)\right\}
\end{equation}
where the regularization is set as $\mathcal{R}_m(\x)=\frac{1}{\eta}\|\x-\y_1\|_2^2$ for convex functions \citep{Hazan2016} and $\mathcal{R}_m(\x)=\sum_{i=1}^{m}\frac{K\alpha}{2}\|\x-\y_i\|_2^2$ for $\alpha$-strongly convex functions \citep{Hazan_2007}.~Unfortunately, due to the effect of delays, the value of $\sum_{i=1}^{m}\nabla_i$ required by \eqref{Ideal-FTRL} may not be available. To address this limitation, we replace it with the sum of all available estimated gradients, i.e., $\bar{\g}_{mK}$.\footnote{From the above discussions, one may replace $\sum_{i=1}^{m}\nabla_i$ with the sum of all available $\nabla_i$. However, we find that simply utilizing $\bar{\g}_{mK}$ can attain the same regret, though they have a slight difference.} 

The detailed procedures are outlined in Algorithm \ref{alg1}, where the input $\alpha$ is the modules of the strong convexity of functions, and this algorithm is called delayed follow-the-bandit-leader (D-FTBL).

\subsection{Theoretical guarantees}
\label{sec-TG}
We first present the regret bound of our D-FTBL for convex functions.
\begin{thm}
\label{thm1}
Under Assumptions \ref{assum3}, \ref{assum1}, and \ref{assum2}, Algorithm \ref{alg1} with $\alpha=0$ ensures
\begin{align*}
\E\left[\re(T)\right]\leq&\frac{4R^2}{\eta}+\frac{\eta T\gamma}{2K}+\frac{\eta TG}{2}\sqrt{2\left(\frac{d^2}{K^2}+4\right)\gamma}+3\delta GT+\frac{\delta GRT}{r}
\end{align*}
where $\gamma=K\left(\frac{nM}{\delta}\right)^2+K^2G^2$.
\end{thm}
\textbf{Remark.} From Theorem \ref{thm1}, by setting $\alpha=0$, $K=n\sqrt{T}$, $\eta=1/\max\{\sqrt{Td},\sqrt{n}T^{3/4}\}$, and $\delta=c\sqrt{n}T^{-1/4}$, where $c$ is a constant such that $\delta<r$, our D-FTBL can achieve the following regret bound 
\begin{equation}
\label{cor-1}
\begin{split}
\E\left[\re(T)\right]\leq&O\left(\sqrt{n}T^{3/4}+\sqrt{dT}\right)
\end{split}
\end{equation}
for convex functions. It is tighter than the $O(\sqrt{n}T^{3/4}+(nd)^{1/3}T^{2/3})$ regret of GOLD \citep{ICML20_Mertikopoulos}, and matches the $O(\sqrt{n}T^{3/4})$ regret bound of BGD in the non-delayed setting as long as $d$ is not larger than $O(n\sqrt{T})$. Even for $d=\Omega(n\sqrt{T})$, our regret bound is dominated by the $O(\sqrt{dT})$ part, which matches the $\Omega(\sqrt{\bar{d}T})$ lower bound \citep{Bistritz-JMLR22} in the worst case.~Moreover, although the $O(\sqrt{n}T^{3/4}+(n\bar{d})^{1/3}T^{2/3})$ regret bound of \citet{Bistritz-JMLR22} could benefit from a small average delay, it is also worse than our regret bound when $d$ is not larger than $O((n\bar{d})^{2/3}T^{1/3})$.

\textbf{Remark.}
One may notice that the step size for achieving the regret bound in \eqref{cor-1} depends on the maximum delay $d$, which may be unknown beforehand. Fortunately, as discussed in previous studies \citep{Quanrud15,Wan-Arxiv-22}, there exists a standard solution---utilizing the ``doubling trick'' \citep{LEA97} to adaptively estimate the maximum delay $d$ and adjust the step size, which can attain the same bound as in \eqref{cor-1}.


Then, we establish an improved regret bound for $\alpha$-strongly convex functions.
\begin{thm}
\label{thm2}
Under Assumptions \ref{assum3}, \ref{assum1}, and \ref{assum2}, if all functions are $\alpha$-strongly convex, Algorithm \ref{alg1} with $\alpha>0$ ensures
\begin{equation*}
\begin{split}
\E\left[\re(T)\right]\leq&(1+\ln T)\left(\frac{2\gamma}{\alpha K}+\frac{G}{\alpha}\sqrt{2\left(\frac{d^2}{K^2}+4\right)\gamma}\right)+(6+4\ln T)R\sqrt{\gamma}+3\delta GT+\frac{\delta GRT}{r}
\end{split}
\end{equation*}
where $\gamma=K\left(\frac{nM}{\delta}\right)^2+K^2G^2$.
\end{thm}
\textbf{Remark.} From Theorem \ref{thm2}, by setting $\alpha>0$, $K=(nT)^{2/3}\ln^{-2/3} T$, and $\delta=cn^{2/3}T^{-1/3}\ln^{1/3}T$, where $c$ is a constant such that $\delta<r$, our D-FTBL enjoys
\begin{equation}
\label{cor-2}
\begin{split}
\E\left[\re(T)\right]\leq&O\left((nT)^{2/3}\log^{1/3} T+d\log T\right)
\end{split}
\end{equation}
for strongly convex functions, which is tighter than the above $O(\sqrt{n}T^{3/4}+\sqrt{dT})$ regret bound achieved by only utilizing the convexity condition. Moreover, it matches the ${O}((nT)^{2/3}\log^{1/3} T)$ regret bound of BGD in the non-delayed setting as long as $d$ is not larger than $O((nT/\log T)^{2/3})$. Even if $d=\Omega((nT/\log T)^{2/3})$, this bound is dominated by the $O(d\log T)$ part, which matches the $\Omega(d\log T)$ lower bound \citep{Weinberger02_TIT}, and thus cannot be improved. Finally, different from the case with convex functions, the parameters for achieving the bound in \eqref{cor-2} do not require the information of delays.

Furthermore, we consider the unconstrained case, i.e., $\K=\mathbb{R}^n$, with $\alpha$-strongly convex and $\beta$-smooth functions, and extend our D-FTBL to achieve a better regret bound. Specifically, without the boundedness of $\K$, Assumptions \ref{assum1} and \ref{assum2} may no longer hold over the entire space \citep{Agarwal2010_COLT}. Therefore, we first introduce a weaker assumption on the Lipschitz continuity, i.e, all loss functions are $G$-Lipschitz at $\ze$. Combining with \eqref{strong-nice-pro}, it is not hard to verify that the fixed optimal action $\x^\ast=\argmin_{\x\in\mathbb{R}^n}\sum_{t=1}^Tf_t(\x)$ satisfies
\begin{equation}
\label{unbound2bound}
\|\x^\ast\|_2\leq\frac{2G}{\alpha}.
\end{equation}
As a result, the player only needs to select actions from the following set
\begin{equation}
\label{set-11}
\K^\prime=\left\{\x\in\mathbb{R}^n\left|\|\x\|_2\leq\frac{2G}{\alpha}\right.\right\}
\end{equation}
which satisfies Assumption \ref{assum3} with $r=R=2G/\alpha$, and it is natural to further assume that all loss functions satisfy Assumptions \ref{assum1} and \ref{assum2} over the set $\K^\prime$. 

Now, we can apply our D-FTBL over the shrink set of $\K^\prime$, i.e.,
\begin{equation}
\label{unconstocons}
\K_\delta^\prime=(1-\delta/r)\K^\prime=\left(1-\frac{\alpha\delta}{2G}\right)\K^\prime
\end{equation}
instead of the original $\K_\delta$, and establish the following regret bound.
\begin{thm}
\label{thm2-1}
Let $\K=\mathbb{R}^n$. If all loss functions are $\alpha$-strongly convex and $\beta$-smooth over $\K$, and Assumptions \ref{assum1} and \ref{assum2} hold over $\K^\prime$ defined in \eqref{set-11}, applying Algorithm \ref{alg1} with $\alpha>0$ over $\K_\delta^\prime$ defined in \eqref{unconstocons} ensures
\begin{equation*}
\begin{split}
\E\left[\re(T)\right]\leq&(1+\ln T)\left(\frac{2\gamma}{\alpha K}+\frac{G}{\alpha}\sqrt{2\left(\frac{d^2}{K^2}+4\right)\gamma}\right)+(6+4\ln T)\frac{2G\sqrt{\gamma}}{\alpha}+\beta\delta^2T+\frac{\beta\delta^2G T}{\alpha}
\end{split}
\end{equation*}
where $\gamma=K\left(\frac{nM}{\delta}\right)^2+K^2G^2$.
\end{thm}
\textbf{Remark.} From Theorem \ref{thm2-1}, by setting $\alpha>0$, $K=n\sqrt{T/\ln T}$, and $\delta=cn^{1/2}T^{-1/4}\ln^{1/4}T$, where $c$ is a constant such that $\delta<2G/\alpha$, in the unconstrained case, we can achieve a regret bound of $O\left(n\sqrt{T\log T}+d\log T\right)$ for strongly convex and smooth functions. It is better than the $O((nT)^{2/3}\log^{1/3} T+d\log T)$ regret bound achieved by only utilizing the strong convexity. Moreover, it matches the $O(n\sqrt{T\log T})$ regret bound achieved by using BGD in the non-delayed setting as long as $d$ is not larger than $O(n\sqrt{T/\log T})$. Otherwise, this bound is dominated by the $O(d\log T)$ part, which cannot be improved as discussed before.




\section{Analysis}
\label{sec-Ana}

We provide the proof of Theorem \ref{thm1} in this section, and the omitted proofs can be found in the appendix.

\subsection{Proof of Theorem \ref{thm1}}
\label{sec-4.1}
We start this proof by introducing the following lemmas.
\begin{lem} 
\label{pre_thm1}
Let $\tilde{\x}^\ast=(1-\delta/r)\x^\ast$, where $\x^\ast\in\argmin_{\x\in\K}\sum_{t=1}^T f_t(\x)$. For each block $m\in[T/K]$, let $\y_m^\ast=\argmin_{\x\in\K_\delta}\left\{\sum_{i=1}^{m-1}\left\langle\nabla_i,\x\right\rangle+\frac{1}{\eta}\|\x-\y_1\|_2^2\right\}$. Then, under Assumptions \ref{assum3} and \ref{assum1}, Algorithm \ref{alg1} with $\alpha=0$ ensures
\begin{equation}
\label{thm1-eq2-1-nw}
\begin{split}
\E\left[\re(T)\right]\leq\E\left[\sum_{m=1}^{T/K}\left\langle\nabla_m, \y_m^\ast-\tilde{\x}^\ast\right\rangle+KG\sum_{m=1}^{T/K}\|\y_m-\y_m^\ast\|_2\right]+3\delta GT+\frac{\delta GRT}{r}.
\end{split}
\end{equation}
\end{lem}
\begin{lem}
\label{lem-ftl}
(Lemma 6.6 in \citet{Garber16}) Let $\{\ell_t(\x)\}_{t=1}^T$ be a sequence of functions over a set $\K$, and let $\x_t^\ast\in\argmin_{\x\in\K}\sum_{i=1}^t\ell_{i}(\x)$ for any $t\in[T]$. Then, it holds that $\sum_{t=1}^T\ell_t(\x_t^\ast)-\min_{\x\in\K}\sum_{t=1}^T\ell_t(\x)\leq 0$.
\end{lem}
\begin{lem}
\label{lem-stab}
(Lemma 5 in \citet{DADO2011}) Let $\Pi_\K(\u,\eta)=\argmin_{\x\in\K}\left\{\langle\u,\x\rangle+\frac{1}{\eta}\|\x\|_2^2\right\}$. We have $\|\Pi_\K(\u,\eta)-\Pi_\K(\v,\eta)\|_2\leq\frac{\eta}{2}\|\u-\v\|_2$.
\end{lem}
To apply Lemma \ref{lem-ftl}, we define $\ell_1(\x)=\langle\nabla_1,\x\rangle+\frac{1}{\eta}\|\x-\y_1\|_2$, and $\ell_m(\x)=\langle\nabla_m,\x\rangle$ for any $m=2,\dots,T/K$. Following the notations in Lemma \ref{pre_thm1}, it is easy to verify that
\begin{equation*}
\begin{split}
    &\sum_{m=1}^{T/K}\left\langle\nabla_m, \y_{m+1}^\ast-\tilde{\x}^\ast\right\rangle+\frac{\|\y_2^\ast-\y_1\|_2^2}{\eta}-\frac{\|\tilde{\x}^\ast-\y_1\|_2^2}{\eta}=\sum_{m=1}^{T/K}\ell_m(\y_{m+1}^\ast)-\sum_{m=1}^{T/K}\ell_m(\tilde{\x}^\ast)\leq0
\end{split}
\end{equation*}
where the inequality is derived by applying Lemma \ref{lem-ftl} to functions $\{\ell_m(\x)\}_{m=1}^{T/K}$ and the set $\K_\delta$.

From the above inequality, we further have
\begin{equation}
\label{thm1-eq4-nw}
\begin{split}
    \sum_{m=1}^{T/K}\left\langle\nabla_m, \y_m^\ast-\tilde{\x}^\ast\right\rangle=&\sum_{m=1}^{T/K}\left\langle\nabla_m, \y_{m+1}^\ast-\tilde{\x}^\ast+\y_{m}^\ast-\y_{m+1}^\ast\right\rangle\\
    \leq&\frac{\|\tilde{\x}^\ast-\y_1\|_2^2}{\eta}-\frac{\|\y_2^\ast-\y_1\|_2^2}{\eta}+\sum_{m=1}^{T/K}\|\nabla_m\|_2 \|\y_{m}^\ast-\y_{m+1}^\ast\|_2\\
    \leq&\frac{4R^2}{\eta}+\frac{\eta}{2}\sum_{m=1}^{T/K}\|\nabla_m\|_2^2
\end{split}
\end{equation}
where the last inequality is due to Assumption \ref{assum3} and Lemma \ref{lem-stab}, i.e.,
\begin{align*}
\|\y_{m}^\ast-\y_{m+1}^\ast\|_2\leq\frac{\eta}{2}\left\|\left(\sum_{i=1}^{m-1}\nabla_i-\frac{2\y_1}{\eta}\right)-\left(\sum_{i=1}^{m}\nabla_i-\frac{2\y_1}{\eta}\right)\right\|_2=\frac{\eta}{2}\left\|\nabla_m\right\|_2.
\end{align*}

We notice that the term $\left\|\nabla_m\right\|_2^2$ in \eqref{thm1-eq4-nw} can directly benefit from the blocking update mechanism, as shown by the upper bound in the following lemma.
\begin{lem}
\label{block-gradient}
Under Assumptions \ref{assum1} and \ref{assum2}, for any $m\in[T/K]$, Algorithm \ref{alg1} ensures
$\E[\|\nabla_m\|_2^2]\leq K\left(\frac{nM}{\delta}\right)^2+K^2G^2$.
\end{lem}
However, to completely bound the right side of \eqref{thm1-eq2-1-nw}, we still need to analyze $\|\y_m-\y_m^\ast\|_2$, which is more complicated due to the effect of delays. Specifically, let 
\begin{equation}
\label{un-rec}
\mathcal{U}_m=\{1,\dots,(m-1)K\}\setminus\cup_{t=1}^{(m-1)K}\F_t
\end{equation} be the set consisting of the time stamp of loss values that are queried but still not arrive at the end of round $(m-1)K$. By using Lemma \ref{lem-stab} again, we have
\begin{equation}
\label{thm1-eq5}
\begin{split}
    \|\y_m-\y_m^\ast\|_2
    \leq\frac{\eta}{2}\left\|\left(\bar{\g}_{(m-1)K}-\frac{2\y_1}{\eta}\right)-\left(\sum_{i=1}^{m-1}\nabla_i-\frac{2\y_1}{\eta}\right)\right\|_2=\frac{\eta}{2}\left\|\sum_{t\in\mathcal{U}_m}\frac{n}{\delta}f_t(\x_t)\u_t\right\|_2.
\end{split}
\end{equation}
Moreover, we establish the following lemma regarding the right side of \eqref{thm1-eq5}.
\begin{lem}
\label{delay-block-gradient}
Under Assumptions \ref{assum1} and \ref{assum2}, for any $m\in[T/K]$, Algorithm \ref{alg1} ensures
\begin{align*}
\E\left[\left\|\sum_{t\in\mathcal{U}_m}\frac{n}{\delta}f_t(\x_t)\u_t\right\|_2^2\right]\leq 2\left(\frac{d^2}{K^2}+4\right)\left(K\left(\frac{nM}{\delta}\right)^2+K^2G^2\right)
\end{align*}
where $\mathcal{U}_m$ is defined in \eqref{un-rec}.
\end{lem}
Let $\gamma=K\left(\frac{nM}{\delta}\right)^2+K^2G^2$. Combining \eqref{thm1-eq2-1-nw}, \eqref{thm1-eq4-nw}, \eqref{thm1-eq5}, and Lemmas \ref{block-gradient} and \ref{delay-block-gradient}, we have
\begin{equation*}
\begin{split}
\E\left[\re(T)\right]-\left(3\delta GT+\frac{\delta GRT}{r}\right)\leq&\E\left[\sum_{m=1}^{T/K}\left\langle\nabla_m, \y_m^\ast-\tilde{\x}^\ast\right\rangle\right]+KG\sum_{m=1}^{T/K}\E\left[\|\y_m-\y_m^\ast\|_2\right]\\
\leq&\frac{4R^2}{\eta}+\E\left[\frac{\eta}{2}\sum_{m=1}^{T/K}\|\nabla_m\|_2^2\right]+KG\sum_{m=1}^{T/K}\E\left[\|\y_m-\y_m^\ast\|_2\right]\\
    \leq&\frac{4R^2}{\eta}+\frac{\eta T\gamma}{2K}+\frac{\eta TG}{2}\sqrt{2\left(\frac{d^2}{K^2}+4\right)\gamma}.
\end{split}
\end{equation*}




\section{Conclusion and future work}
\label{Con}
In this paper, we investigate BCO with delayed feedback, and propose a novel algorithm called D-FTBL by exploiting the blocking update
mechanism. Our analysis first reveals that it can achieve a regret bound of $O(\sqrt{n}T^{3/4}+\sqrt{dT})$ in general, which improves the delay-dependent part of the existing $O(\sqrt{n}T^{3/4}+(n\bar{d})^{1/3}T^{2/3})$ regret bound as long as $d$ is not larger than $O((n\bar{d})^{2/3}T^{1/3})$. Furthermore, we consider the special case with strongly convex functions, and prove that the regret of D-FTBL can be reduced to $O((nT)^{2/3}\log^{1/3}T+d\log T)$. Finally, if the action sets are unconstrained, we show that D-FTBL can be simply extended to enjoy the $O(n\sqrt{T\log T}+d\log T)$ regret for strongly convex and smooth functions.

Note that all our regret bounds depend on the maximum delay. A natural open problem is whether these bounds can be further improved to be depending on the average delay. It seems highly non-trivial to obtain such results with our D-FTBL because the blocking update mechanism actually enlarges each delay to be at least the block size, and thus we leave this problem as a future work. Additionally, it is also appealing to extend other BCO algorithms into the delayed setting, e.g., generalizing the algorithm of \citet{Saha_2011} to keep the $O((nT)^{2/3}\log^{1/3}T)$ regret for smooth functions under a certain amount of delay. However, they generally utilize additional techniques, e.g., the self-concordant barrier \citep{Nemirovski-04-Lec}, which require a more complicated analysis.

\newpage 
\bibliographystyle{plainnat}
\bibliography{ref}

\newpage
\appendix

\section{Proof of Lemma \ref{pre_thm1}}
Let $\mathcal{T}_m=\{(m-1)K+1,\dots,mK\}$ for brevity. We first notice that
\begin{equation}
\label{thm1-pre-eq1}
\begin{split}
\re(T)=&\sum_{m=1}^{T/K}\sum_{t\in\mathcal{T}_m}(f_t(\y_m+\delta\u_t)-f_t(\x^\ast))\\
\leq&\sum_{m=1}^{T/K}\sum_{t\in\mathcal{T}_m}\left(f_t(\y_m)+G\|\delta\u_t\|_2\right)-\sum_{m=1}^{T/K}\sum_{t\in\mathcal{T}_m}\left(f_t(\tilde{\x}^\ast)-G\left\|\frac{\delta}{r}\x^\ast\right\|_2\right)\\
\leq&\sum_{m=1}^{T/K}\sum_{t\in\mathcal{T}_m}\left(f_t(\y_m)-f_t(\tilde{\x}^\ast)\right)+\delta GT+\frac{\delta GRT}{r}
\end{split}
\end{equation}
where the first inequality is due to Assumption \ref{assum1}, and the second inequality is due to Assumption \ref{assum3} and $\u_t\sim\mathcal{S}^n$.

According to Algorithm \ref{alg1}, $\y_1,\dots,\y_{T/K}$ are computed according to approximate gradients of the $\delta$-smoothed version of original loss functions. Therefore, we introduce the following lemma regarding the $\delta$-smoothed function.
\begin{lem}
\label{smoothed_lem1}
(Lemma 2.6 in \citet{Hazan2016})
Let $f(\x):\mathbb{R}^n\to\mathbb{R}$ be $\alpha$-strongly convex and $G$-Lipschitz over a set $\K$ satisfying Assumption \ref{assum3}. Its $\delta$-smoothed version $\hat{f}_\delta(\x)$ defined in \eqref{delta-smooth} has the following properties:
\begin{compactitem}
\item $\hat{f}_\delta(\x)$ is $\alpha$-strongly convex over $\K_\delta$;
\item $|\hat{f}_\delta(\x)-f(\x)|\leq\delta G$ for any $\x\in\K_\delta$;
\item $\hat{f}_\delta(\x)$ is $G$-Lipschitz over $\K_\delta$;
\end{compactitem}
where $\K_\delta$ is defined in \eqref{shrink-set}.
\end{lem}
Combining \eqref{thm1-pre-eq1} with the second property in Lemma \ref{smoothed_lem1}, it is easy to verify that
\begin{equation}
\label{thm1-eq1}
\begin{split}
\re(T)\leq&\sum_{m=1}^{T/K}\sum_{t=(m-1)K+1}^{mK}\left(\hat{f}_{t,\delta}(\y_m)-\hat{f}_{t,\delta}(\tilde{\x}^\ast)+2\delta G\right)+\delta GT+\frac{\delta GRT}{r}\\
\leq&\sum_{m=1}^{T/K}\sum_{t=(m-1)K+1}^{mK}\langle\nabla \hat{f}_{t,\delta}(\y_m), \y_m-\tilde{\x}^\ast\rangle+3\delta GT+\frac{\delta GRT}{r}
\end{split}
\end{equation}
where the last inequality is due to the convexity of loss functions.

From \eqref{thm1-eq1}, we can further have
\begin{equation}
\label{thm1-eq2-1}
\begin{split}
&\re(T)-\left(3\delta GT+\frac{\delta GRT}{r}\right)\\
\leq&\sum_{m=1}^{T/K}\sum_{t\in\mathcal{T}_m}\langle\nabla \hat{f}_{t,\delta}(\y_m), \y_m^\ast-\tilde{\x}^\ast+\y_m-\y_m^\ast\rangle\\
\leq&\sum_{m=1}^{T/K}\sum_{t\in\mathcal{T}_m}\langle\nabla \hat{f}_{t,\delta}(\y_m), \y_m^\ast-\tilde{\x}^\ast\rangle+KG\sum_{m=1}^{T/K}\|\y_m-\y_m^\ast\|_2\\
\end{split}
\end{equation}
where the last inequality is due to Assumption \ref{assum1} and the last property in Lemma \ref{smoothed_lem1}.

Moreover, according to Lemma \ref{smoothed_lem2}, we have
\begin{equation}
\label{thm1-eq2}
\begin{split}
\E\left[\sum_{m=1}^{T/K}\sum_{t\in\mathcal{T}_m}\langle\nabla \hat{f}_{t,\delta}(\y_m), \y_m^\ast-\tilde{\x}^\ast\rangle\right]=&\E\left[\sum_{m=1}^{T/K}\sum_{t\in\mathcal{T}_m}\left\langle\frac{n}{\delta}f_t(\y_m+\delta\u_t)\u_t, \y_m^\ast-\tilde{\x}^\ast\right\rangle\right]\\
=&\E\left[\sum_{m=1}^{T/K}\left\langle\nabla_m, \y_m^\ast-\tilde{\x}^\ast\right\rangle\right].
\end{split}
\end{equation}
Combining \eqref{thm1-eq2-1} with \eqref{thm1-eq2}, this proof is completed.

\section{Proof of Lemmas \ref{block-gradient} and \ref{delay-block-gradient}}
Lemma \ref{block-gradient} can be proved by simply following the proof of Lemma 5 in \citet{Garber19}. In the following, we first prove Lemma \ref{delay-block-gradient}, and then include a simple proof of Lemma \ref{block-gradient} for completeness.

For brevity, let $\g_t=\frac{n}{\delta}f_t(\x_t)\u_t$ for any $t\in[T]$. Since $\g_1,\dots,\g_{(m-1)K-d+1}$ must be available at the end of round $(m-1)K$, it is not hard to verify that
\begin{equation}
\label{lem1-eq1}
\begin{split}
\left\|\sum_{t\in\mathcal{U}_m}\g_t\right\|_2^2=&\left\|\sum_{k=m-1-\lceil d/K\rceil}^{m-1}\sum_{t\in \mathcal{T}_k\cap\mathcal{U}_m}\g_t\right\|_2^2\leq\left(\left\lceil \frac{d}{K}\right\rceil+1\right)\sum_{k=m-1-\lceil d/K\rceil}^{m-1}\left\|\sum_{t\in \mathcal{T}_k\cap\mathcal{U}_m}\g_t\right\|_2^2
\end{split}
\end{equation}
where $\mathcal{T}_k=\{(k-1)K+1,\dots,kK\}$. 

Moreover, let $\A_k=\mathcal{T}_k\cap\mathcal{U}_m$. It is easy to verify that $|\A_k|\leq|\mathcal{T}_k|=K$.~Then, for any $k=m-1-\lceil d/K\rceil,\dots,m-1$, we have
\begin{equation}
\label{lem1-eq2}
\begin{split}
\E\left[\left\|\sum_{t\in \mathcal{T}_k\cap\mathcal{U}_m}\g_t\right\|_2^2\right]=&\E\left[\sum_{t\in \A_k}\left\|\g_t\right\|_2^2+\sum_{i,j\in \A_k,i\neq j }\langle\g_i,\g_j\rangle\right]\\
\leq&|\A_k|\left(\frac{nM}{\delta}\right)^2+\E\left[\sum_{i,j\in \A_k,i\neq j }\left\langle\E[\g_i|\y_{k}],\E[\g_j|\y_{k}]\right\rangle\right]\\
\leq&K\left(\frac{nM}{\delta}\right)^2+\E\left[\sum_{i,j\in \A_k,i\neq j }\|\E[\g_i|\y_{k}]\|_2\|\E[\g_j|\y_{k}]\|_2\right]\\
\leq &K\left(\frac{nM}{\delta}\right)^2+(|\A_k|^2-|\A_k|)G^2\leq K\left(\frac{nM}{\delta}\right)^2+K^2G^2
\end{split}
\end{equation}
where the first inequality is due to Assumption \ref{assum2}, and the third inequality is due to Assumption \ref{assum1}, Lemma \ref{smoothed_lem2}, and the last property in Lemma \ref{smoothed_lem1}.

Combining \eqref{lem1-eq1} with \eqref{lem1-eq2}, we have 
\begin{equation*}
\begin{split}
\E\left[\left\|\sum_{t\in\mathcal{U}_m}\g_t\right\|_2^2\right]\leq 2\left(\frac{d^2}{K^2}+4\right)\left(K\left(\frac{nM}{\delta}\right)^2+K^2G^2\right)
\end{split}
\end{equation*}
which completes the proof of Lemma \ref{delay-block-gradient}. 

Additionally, following \eqref{lem1-eq2}, it is also not hard to verify that
\begin{equation}
\label{lem1-eq2-222}
\begin{split}
\E\left[\left\|\nabla_m\right\|_2^2\right]
=\E\left[\sum_{t\in \mathcal{T}_m}\left\|\g_t\right\|_2^2+\sum_{i,j\in \mathcal{T}_m,i\neq j }\langle\g_i,\g_j\rangle\right]\leq K\left(\frac{nM}{\delta}\right)^2+K^2G^2
\end{split}
\end{equation}
which completes the proof of Lemma \ref{block-gradient}.

\section{Proof of Theorem \ref{thm2}}
This proof is similar to that of Theorem \ref{thm1}, but requires some specific extensions to utilize the strong convexity. 
Specifically, we define a sequence of functions, i.e., $\ell_m(\x)=\langle\nabla_m, \x\rangle+\frac{\alpha K}{2}\|\y_m-\x\|_2^2$ for $m\in[T/K]$, and an ideal decision
\begin{equation}
\label{thm2-eq2}
\y_m^\ast=\argmin_{\x\in\K_\delta}\sum_{i=1}^{m-1}\ell_i(\x)=\argmin_{\x\in\K_\delta}\left\{\left\langle\sum_{i=1}^{m-1}\nabla_i,\x\right\rangle+\sum_{i=1}^{m-1}\frac{\alpha K}{2}\|\y_i-\x\|_2^2\right\}
\end{equation}
for each block $m=2,\dots,T/K$. 

Then, combining  \eqref{thm1-eq1} in the proof of Lemma \ref{pre_thm1} with the strong convexity of functions, we have
\begin{equation}
\label{thm2-eq1}
\begin{split}
&\re(T)-\left(3\delta GT+\frac{\delta GRT}{r}\right)\\
\leq&\sum_{m=1}^{T/K}\sum_{t\in\mathcal{T}_m}\left(\langle\nabla \hat{f}_{t,\delta}(\y_m), \y_m-\tilde{\x}^\ast\rangle-\frac{\alpha}{2}\|\y_m-\tilde{\x}^\ast\|_2^2\right)\\
=&\sum_{m=1}^{T/K}\sum_{t\in\mathcal{T}_m}\left(\langle\nabla \hat{f}_{t,\delta}(\y_m), \y_m^\ast-\tilde{\x}^\ast\rangle+\langle\nabla \hat{f}_{t,\delta}(\y_m), \y_m-\y_m^\ast\rangle-\frac{\alpha}{2}\|\y_m-\tilde{\x}^\ast\|_2^2\right)\\
\leq&\sum_{m=1}^{T/K}\sum_{t\in\mathcal{T}_m}\left(\langle\nabla \hat{f}_{t,\delta}(\y_m), \y_m^\ast-\tilde{\x}^\ast\rangle-\frac{\alpha}{2}\|\y_m-\tilde{\x}^\ast\|_2^2\right)+\sum_{m=1}^{T/K}KG\|\y_m-\y_m^\ast\|_2
\end{split}
\end{equation}
where we simply set $\y_1^\ast=\y_1$, and the last inequality is due to Assumption \ref{assum1} and the last property in Lemma \ref{smoothed_lem1}.

Moreover, it is not hard to verify that
\begin{equation}
\label{thm2-eq3}
\begin{split}
&\E\left[ \sum_{m=1}^{T/K}\sum_{t\in\mathcal{T}_m}\left(\langle\nabla \hat{f}_{t,\delta}(\y_m), \y_m^\ast-\tilde{\x}^\ast\rangle-\frac{\alpha}{2}\|\y_m-\tilde{\x}^\ast\|_2^2\right)\right]\\
=&\E\left[ \sum_{m=1}^{T/K}\sum_{t\in\mathcal{T}_m}\left(\left\langle\frac{n}{\delta}f_t(\y_m+\delta\u_t)\u_t, \y_m^\ast-\tilde{\x}^\ast\right\rangle-\frac{\alpha}{2}\|\y_m-\tilde{\x}^\ast\|_2^2\right)\right]\\
=&\E\left[ \sum_{m=1}^{T/K}\left(\left\langle\nabla_m, \y_{m+1}^\ast-\tilde{\x}^\ast+\y_m^\ast-\y_{m+1}^\ast\right\rangle-\frac{\alpha K}{2}\|\y_m-\tilde{\x}^\ast\|_2^2\right)\right]\\
\leq&\E\left[ \sum_{m=1}^{T/K}\left(\ell_m(\y_{m+1}^\ast)-\ell_m(\tilde{\x}^\ast)\right)\right]+\E\left[\sum_{m=1}^{T/K}\|\nabla_m\|_2\|\y_m^\ast-\y_{m+1}^\ast\|_2\right]\\
\leq&\E\left[\sum_{m=1}^{T/K}\|\nabla_m\|_2\|\y_m^\ast-\y_{m+1}^\ast\|_2\right]
\end{split}
\end{equation}
where the first equality is due to Lemma \ref{smoothed_lem2}, and the last inequality is due to Lemma \ref{lem-ftl}.

From \eqref{thm2-eq1} and \eqref{thm2-eq3}, we still need to bound $\|\y_m-\y_m^\ast\|_2$ and $\|\y_m^{\ast}-\y_{m+1}^\ast\|_2$. To this end, we notice that $\y_m^\ast$ defined in \eqref{thm2-eq2} is equal to
\begin{equation}
\label{thm2-eq4}
\begin{split}
\y_m^\ast=&\argmin_{\x\in\K_\delta}\left\{\left\langle\sum_{i=1}^{m-1}(\nabla_i-\alpha K\y_i),\x\right\rangle+\frac{\alpha (m-1)K}{2}\|\x\|_2^2\right\}.
\end{split}
\end{equation}
Similarly, for any $m=2,\dots,T/K$, the decision $\y_{m}$ of Algorithm \ref{alg1} with $\alpha>0$ is equal to
\begin{equation}
\label{thm2-eq5}
\y_{m}=\argmin_{\x\in\K_\delta}\left\{\left\langle\bar{\g}_{(m-1)K}-\sum_{i=1}^{m-1}\alpha K\y_i,\x\right\rangle+\frac{\alpha (m-1)K}{2}\|\x\|_2^2\right\}.
\end{equation}
Combining \eqref{thm2-eq4} and \eqref{thm2-eq5} with  Lemma \ref{lem-stab}, for any $m=2,\dots,T/K$, we have
\begin{equation}
\label{thm2-eq6}
\begin{split}
    \|\y_{m}-\y_{m}^\ast\|_2\leq\frac{1}{\alpha (m-1)K}\left\|\bar{\g}_{(m-1)K}-\sum_{i=1}^{m-1}\nabla_i\right\|_2=\frac{1}{\alpha (m-1)K}\left\|\sum_{t\in\mathcal{U}_{m}}\frac{n}{\delta}f_t(\x_t)\u_t\right\|_2
\end{split}
\end{equation}
where $\mathcal{U}_m$ is defined in \eqref{un-rec}.

Moreover, from \eqref{strong-nice-pro}, for any $m=2,\dots,T/K$, we have
\begin{equation*}
\begin{split}
    \|\y_{m}^\ast-\y_{m+1}^\ast\|_2^2\leq&\frac{2}{\alpha mK}\left(\sum_{i=1}^{m}\ell_i(\y_{m}^\ast)-\sum_{i=1}^{m}\ell_i(\y_{m+1}^\ast)\right)\leq\frac{2}{\alpha mK}\left(\ell_m(\y_{m}^\ast)-\ell_m(\y_{m+1}^\ast)\right)\\
     \leq&\frac{2}{\alpha mK}\left(\langle\nabla_m+\alpha K(\y_{m}^\ast-\y_m),\y_{m}^\ast-\y_{m+1}^\ast\rangle\right)\\
     \leq&\frac{2}{\alpha (m-1)K}\left(\|\nabla_m\|_2+2\alpha KR\right)\|\y_{m}^\ast-\y_{m+1}^\ast\|_2
\end{split}
\end{equation*}
where the last inequality is due to Assumption \ref{assum3}. The above inequality further implies that
\begin{equation}
\label{thm2-eq7}
\begin{split}
    \|\y_{m}^\ast-\y_{m+1}^\ast\|_2\leq\frac{2}{\alpha (m-1)K}\left(\|\nabla_m\|_2+2\alpha KR\right).
\end{split}
\end{equation}
Combining \eqref{thm2-eq1}, \eqref{thm2-eq3}, \eqref{thm2-eq6}, and \eqref{thm2-eq7}, we have
\begin{equation*}
\begin{split}
&\E\left[\re(T)\right]-\left(3\delta GT+\frac{\delta GRT}{r}\right)\\
\leq&\E\left[\sum_{m=1}^{T/K}\|\nabla_m\|_2\|\y_m^\ast-\y_{m+1}^\ast\|_2\right]+\sum_{m=1}^{T/K}KG\E\left[\|\y_m-\y_m^\ast\|_2\right]\\
\leq&\E[\|\nabla_1\|_2\|\y_{1}-\y_{2}^\ast\|_2]+\E\left[\sum_{m=2}^{T/K}\frac{2\left(\|\nabla_m\|_2^2+2\alpha KR\|\nabla_m\|_2\right)}{\alpha(m-1)K}\right]\\
&+\frac{G}{\alpha(m-1)}\sum_{m=2}^{T/K}\E\left[\left\|\sum_{t\in\mathcal{U}_{m}}\frac{n}{\delta}f_t(\x_t)\u_t\right\|_2\right].
\end{split}
\end{equation*}
For brevity, let $\gamma=K\left(\frac{nM}{\delta}\right)^2+K^2G^2$. Combining the above inequality with Assumption \ref{assum3}, and Lemmas \ref{block-gradient} and \ref{delay-block-gradient}, we have
\begin{align*}
\E\left[\re(T)\right]
\leq&2R\sqrt{\gamma}+\sum_{m=2}^{T/K}\frac{1}{m-1}\left(\frac{2\gamma}{\alpha K}+4R\sqrt{\gamma}+\frac{G}{\alpha}\sqrt{2\left(\frac{d^2}{K^2}+4\right)\gamma}\right)+3\delta GT+\frac{\delta GRT}{r}\\
\leq&2R\sqrt{\gamma}+(1+\ln T)\left(\frac{2\gamma}{\alpha K}+4R\sqrt{\gamma}+\frac{G}{\alpha}\sqrt{2\left(\frac{d^2}{K^2}+4\right)\gamma}\right)+3\delta GT+\frac{\delta GRT}{r}\\
=&(1+\ln T)\left(\frac{2\gamma}{\alpha K}+\frac{G}{\alpha}\sqrt{2\left(\frac{d^2}{K^2}+4\right)\gamma}\right)+(6+4\ln T)R\sqrt{\gamma}+3\delta GT+\frac{\delta GRT}{r}.
\end{align*}

\section{Proof of Theorem \ref{thm2-1}}
The main idea of this proof is to combine the proof of Theorem \ref{thm2} with an improved property of the $\delta$-smoothed version of smooth functions \citep{Agarwal2010_COLT}. 

Specifically, for any $t\in[T]$ and $\x$, according to the smoothness of functions, we have
\begin{equation}
\label{uncons-eq1}
\begin{split}
\hat{f}_{t,\delta}(\x)
=&\E_{\u\sim\mathcal{B}^n}\left[f_t(\x+\delta\u)\right]\\
\leq&\E_{\u\sim\mathcal{B}^n}\left[f_t(\x)+\langle\nabla f_t(\x),\delta\u\rangle+\frac{\beta\delta^2\|\u\|_2^2}{2}\right]\\
=&f_t(\x)+\frac{\beta\delta^2}{2}
\end{split}
\end{equation}
where the last equality is due to $\E_{\u\sim\mathcal{B}^n}\left[\u\right]=\ze$.

Moreover, due to the convexity of functions, for any $t\in[T]$ and $\x$, we have
\begin{equation}
\label{uncons-eq1-2}
\begin{split}
\hat{f}_{t,\delta}(\x)=&\E_{\u\sim\mathcal{B}^n}\left[f_t(\x+\delta\u)\right]\geq\E_{\u\sim\mathcal{B}^n}\left[f_t(\x)+\langle\nabla f_t(\x),\delta\u\rangle\right]=f_t(\x).
\end{split}
\end{equation}
Then, let $\x^\ast=\argmin_{\x\in\mathbb{R}^n}\sum_{t=1}^Tf_t(\x)$, and $\tilde{\x}^\ast=(1-\delta/r)\x^\ast$, where $r=2G/\alpha$. According to \eqref{unbound2bound}, we have $\x^\ast\in\K^\prime$ and $\tilde{\x}^\ast\in\K_\delta^\prime$, where $\K^\prime$ and $\K_\delta^\prime$ are defined in \eqref{set-11} and \eqref{unconstocons}, respectively. It is not hard to verify that 
\begin{equation}
\label{uncons-eq2}
\begin{split}
\E\left[\re(T)\right]=&\E\left[\sum_{m=1}^{T/K}\sum_{t=(m-1)K+1}^{mK}(f_t(\y_m+\delta\u_t)-f_t(\x^\ast))\right]\\
\leq&\E\left[\sum_{m=1}^{T/K}\sum_{t=(m-1)K+1}^{mK}\left(f_t(\y_m)+\langle\nabla f_t(\y_m),\delta\u_t\rangle+\frac{\beta\delta^2\|\u_t\|^2_2}{2}\right)\right]\\
&+\E\left[\sum_{m=1}^{T/K}\sum_{t=(m-1)K+1}^{mK}\left(-f_t(\tilde{\x}^\ast)+\left\langle\nabla f_t(\x^\ast),-\frac{\delta {\x}^\ast}{r}\right\rangle+\frac{\beta\delta^2\|\x^\ast\|_2^2}{2r}\right)\right]\\
=&\E\left[\sum_{m=1}^{T/K}\sum_{t=(m-1)K+1}^{mK}\left(f_t(\y_m)-f_t(\tilde{\x}^\ast)+\frac{\beta\delta^2}{2}+\frac{\beta\delta^2G}{\alpha}\right)\right]\\
\leq&\E\left[\sum_{m=1}^{T/K}\sum_{t=(m-1)K+1}^{mK}\left(\tilde{f}_{t,\delta}(\y_m)-\tilde{f}_{t,\delta}(\tilde{\x}^\ast)\right)\right]+\beta\delta^2T+\frac{\beta\delta^2G T}{\alpha}
\end{split}
\end{equation}
where the first inequality is due to the smoothness of functions, and the last inequality is due to \eqref{uncons-eq1} and \eqref{uncons-eq1-2}.

Then, we follow the definition of $\y_m^\ast$ in \eqref{thm2-eq2}, but replace $\K_\delta$ utilized in \eqref{thm2-eq2} with $\K_\delta^\prime$. Combining  \eqref{uncons-eq2} with the strong convexity of functions, we have
\begin{equation}
\label{uncons-eq3}
\begin{split}
&\E\left[\re(T)\right]-\left(\beta\delta^2T+\frac{\beta\delta^2G T}{\alpha}\right)\\
\leq&\E\left[\sum_{m=1}^{T/K}\sum_{t\in\mathcal{T}_m}\left(\langle\nabla \hat{f}_{t,\delta}(\y_m), \y_m-\tilde{\x}^\ast\rangle-\frac{\alpha}{2}\|\y_m-\tilde{\x}^\ast\|_2^2\right)\right]\\
=&\E\left[\sum_{m=1}^{T/K}\sum_{t\in\mathcal{T}_m}\left(\langle\nabla \hat{f}_{t,\delta}(\y_m), \y_m^\ast-\tilde{\x}^\ast\rangle+\langle\nabla \hat{f}_{t,\delta}(\y_m), \y_m-\y_m^\ast\rangle-\frac{\alpha}{2}\|\y_m-\tilde{\x}^\ast\|_2^2\right)\right]\\
\leq&\E\left[\sum_{m=1}^{T/K}\sum_{t\in\mathcal{T}_m}\left(\langle\nabla \hat{f}_{t,\delta}(\y_m), \y_m^\ast-\tilde{\x}^\ast\rangle-\frac{\alpha}{2}\|\y_m-\tilde{\x}^\ast\|_2^2\right)\right]+\E\left[\sum_{m=1}^{T/K}KG\|\y_m-\y_m^\ast\|_2\right]
\end{split}
\end{equation}
where we simply set $\y_1^\ast=\y_1$, and the last inequality is due to Assumption \ref{assum1} and the last property in Lemma \ref{smoothed_lem1}.

Let $R=2G/\alpha$ denote the radius of $\K^\prime$. It is not hard to verify that \eqref{thm2-eq3}, \eqref{thm2-eq6}, and \eqref{thm2-eq7} in the proof of Theorem \ref{thm2} still hold here. Therefore,  we have
\begin{equation}
\label{uncons-eq4}
\begin{split}
&\E\left[\re(T)\right]-\left(\beta\delta^2T+\frac{\beta\delta^2G T}{\alpha}\right)\\
\leq&\E\left[\sum_{m=1}^{T/K}\|\nabla_m\|_2\|\y_m^\ast-\y_{m+1}^\ast\|_2\right]+\E\left[\sum_{m=1}^{T/K}KG\|\y_m-\y_m^\ast\|_2\right]\\
\leq&\E[\|\nabla_1\|_2\|\y_1-\y_{2}^\ast\|_2]+\E\left[\sum_{m=2}^{T/K}\frac{2\left(\|\nabla_m\|_2^2+2\alpha KR\|\nabla_m\|_2\right)}{\alpha(m-1)K}\right]\\
&+\frac{G}{\alpha(m-1)}\sum_{m=2}^{T/K}\E\left[\left\|\sum_{t\in\mathcal{U}_{m}}\frac{n}{\delta}f_t(\x_t)\u_t\right\|_2\right]
\end{split}
\end{equation}
where the first inequality is derived by combing \eqref{uncons-eq3} with \eqref{thm2-eq3}, and the second one is due to \eqref{thm2-eq6} and \eqref{thm2-eq7}.

For brevity, let $\gamma=K\left(\frac{nM}{\delta}\right)^2+K^2G^2$. Combining \eqref{uncons-eq4} with Lemmas \ref{block-gradient} and \ref{delay-block-gradient}, we have
\begin{align*}
\E\left[\re(T)\right]
\leq&2R\sqrt{\gamma}+\sum_{m=2}^{T/K}\frac{1}{m-1}\left(\frac{2\gamma}{\alpha K}+4R\sqrt{\gamma}+\frac{G}{\alpha}\sqrt{2\left(\frac{d^2}{K^2}+4\right)\gamma}\right)+\beta\delta^2T+\frac{\beta\delta^2G T}{\alpha}\\
\leq&2R\sqrt{\gamma}+(1+\ln T)\left(\frac{2\gamma}{\alpha K}+4R\sqrt{\gamma}+\frac{G}{\alpha}\sqrt{2\left(\frac{d^2}{K^2}+4\right)\gamma}\right)+\beta\delta^2T+\frac{\beta\delta^2G T}{\alpha}\\
=&(1+\ln T)\left(\frac{2\gamma}{\alpha K}+\frac{G}{\alpha}\sqrt{2\left(\frac{d^2}{K^2}+4\right)\gamma}\right)+(6+4\ln T)\frac{2G\sqrt{\gamma}}{\alpha}+\beta\delta^2T+\frac{\beta\delta^2G T}{\alpha}
\end{align*}
where the last equality is due to $R=2G/\alpha$.

\section{A refined regret bound for \citet{Bistritz-JMLR22}}
\label{app1}
From Theorem 4 of \citet{Bistritz-JMLR22}, their algorithm can achieve the following regret bound
\begin{equation}
\label{thm4-Bistritz}
\mathbb{E}[\re(T)]=O\left(\delta T+\frac{\eta n^2T}{\delta^2}+\frac{1}{\eta}+\frac{n\eta\bar{d}T}{\delta}\right)
\end{equation}
for BCO with delayed feedback, where $\delta>0$ and $\eta>0$ denote the exploration radius and the step size, respectively. Then, by further substituting 
\begin{equation*}
\delta=\max\left\{T^{-1/4},T^{-1/3}\bar{d}^{1/3}\right\}\text{ and } \eta=\min\left\{n^{-1}T^{-3/4},n^{-1/2}T^{-2/3}\bar{d}^{-1/3}\right\}
\end{equation*}
into \eqref{thm4-Bistritz}, \citet{Bistritz-JMLR22} have established the $O(nT^{3/4}+\sqrt{n}\bar{d}^{1/3}T^{2/3})$ regret bound.

However, it is not hard to verify that
\begin{equation}
\label{improve-thm4-Bistritz}
\begin{split}
\min_{\delta>0,\eta>0}O\left(\delta T+\frac{\eta n^2T}{\delta^2}+\frac{1}{\eta}+\frac{n\eta\bar{d}T}{\delta}\right)
=&\min_{\delta>0}O\left(\delta T+\sqrt{\frac{n^2T}{\delta^2}+\frac{n\bar{d}T}{\delta}}\right)
\end{split}
\end{equation}
where the first equality holds with 
\begin{equation}
\label{improve-eta-thm4-Bistritz}
\eta=\left(\frac{n^2T}{\delta^2}+\frac{n\bar{d}T}{\delta}\right)^{-1/2}.
\end{equation}
From \eqref{improve-thm4-Bistritz}, if $n^2T\delta^{-2}\geq n\bar{d}T\delta^{-1}$, we have
\begin{equation}
\label{improve-thm4-Bistritz-2}
\begin{split}
\min_{\delta>0,\eta>0}O\left(\delta T+\frac{\eta n^2T}{\delta^2}+\frac{1}{\eta}+\frac{n\eta\bar{d}T}{\delta}\right)
=\min_{\delta>0}O\left(\delta T+\frac{n\sqrt{T}}{\delta}\right)=O\left(\sqrt{n}T^{3/4}\right)
\end{split}
\end{equation}
where the last equality holds with $\delta=\sqrt{n}T^{-1/4}$. 

Otherwise, combining \eqref{improve-thm4-Bistritz} with $n^2T\delta^{-2}< n\bar{d}T\delta^{-1}$, we have
\begin{equation}
\label{improve-thm4-Bistritz-3}
\begin{split}
\min_{\delta>0,\eta>0}O\left(\delta T+\frac{\eta n^2T}{\delta^2}+\frac{1}{\eta}+\frac{n\eta\bar{d}T}{\delta}\right)
=\min_{\delta>0}O\left(\delta T+\sqrt{\frac{n\bar{d}T}{\delta}}\right)=O\left((n\bar{d})^{1/3}T^{2/3}\right)
\end{split}
\end{equation}
where the last equality holds with $\delta=(n\bar{d})^{1/3}T^{-1/3}$.

Combining \eqref{thm4-Bistritz} with \eqref{improve-eta-thm4-Bistritz}, \eqref{improve-thm4-Bistritz-2}, and \eqref{improve-thm4-Bistritz-3}, we can improve the regret bound of \citet{Bistritz-JMLR22} to
\begin{equation}
\label{our-thm4-Bistritz}
\mathbb{E}[\re(T)]=O\left(\sqrt{n}T^{3/4}+(n\bar{d})^{1/3}T^{2/3}\right)
\end{equation}
by setting $\delta$ and $\eta$ as 
\begin{equation*}
\delta=\max\left\{\sqrt{n}T^{-1/4},(n\bar{d})^{1/3}T^{-1/3}\right\}\text{ and } \eta=\min\left\{n^{-1/2}T^{-3/4},(n\bar{d})^{-1/3}T^{-2/3}\right\}.
\end{equation*}

\end{document}